\documentclass[10pt,twocolumn,letterpaper]{article}

\usepackage[pagenumbers]{cvpr}
\usepackage[dvipsnames,table]{xcolor}

\usepackage{url}
\usepackage{booktabs}
\usepackage{multirow}
\usepackage{graphicx}
\usepackage{wrapfig}
\usepackage{tikz}

\usepackage{amsmath,amsfonts,bm}

\def\eqref#1{equation~\ref{#1}}

\def\1{\bm{1}}

\DeclareMathAlphabet{\mathsfit}{\encodingdefault}{\sfdefault}{m}{sl}
\SetMathAlphabet{\mathsfit}{bold}{\encodingdefault}{\sfdefault}{bx}{n}

\DeclareMathOperator*{\argmin}{arg\,min}

\newcommand{\plora}{\texttt{p}LoRA\;}
\newcommand{\twolora}{\texttt{2}LoRA\;}

\definecolor{cvprblue}{rgb}{0.21,0.49,0.74}
\usepackage[pagebackref,breaklinks,colorlinks,citecolor=cvprblue]{hyperref}

\def\paperID{}
\def\confName{CVPR}
\def\confYear{2024}

\title{Non-Visible Light Data Synthesis and Application: A Case Study \\for Synthetic Aperture Radar Imagery}

\author{
Zichen Tian \quad Zhaozheng Chen \quad Qianru Sun \\
\\
\small  Singapore Management University \\
\small  {\texttt{\{zichen.tian.2023,zzchen.2019\}@phdcs.smu.edu.sg}} \quad  {\texttt{qianrusun@smu.edu.sg}}
}

\begin{document}
\maketitle
\begin{abstract}
    We explore the ``hidden'' ability of large-scale pre-trained image generation models, such as Stable Diffusion and Imagen, in non-visible light domains, taking Synthetic Aperture Radar (SAR) data for a case study. Due to the inherent challenges in capturing satellite data, acquiring ample SAR training samples is infeasible. For instance, for a particular category of \texttt{ship} in the open sea, we can collect only few-shot SAR images which are too limited to derive effective \texttt{ship} recognition models. If large-scale models pre-trained with regular images can be adapted to generating novel SAR images, the problem is solved. In preliminary study, we found that fine-tuning these models with few-shot SAR images is not working, as the models can not capture the two primary differences between SAR and regular images: structure and modality. To address this, we propose a \underline{2}-stage \underline{lo}w-\underline{r}ank \underline{a}daptation method, and we call it \texttt{2}LoRA. In the first stage, the model is adapted using aerial-view regular image data (whose structure matches SAR), followed by the second stage where the base model from the first stage is further adapted using SAR modality data. Particularly in the second stage, we introduce a novel \underline{p}rototype LoRA (\texttt{p}LoRA), as an improved version of \texttt{2}LoRA, to resolve the class imbalance problem in SAR datasets. For evaluation, we employ the resulting generation model to synthesize additional SAR data. This augmentation, when integrated into the training process of SAR classification as well as segmentation models, yields notably improved performance for minor classes\footnote{Code: \href{https://github.com/doem97/gen_sar_plora}{https://github.com/doem97/gen\_sar\_plora}}.
\end{abstract}
    
\section{Introduction}
\label{sec:intro}
\begin{figure*}[t]
    \centering
    \includegraphics[width=1\linewidth]{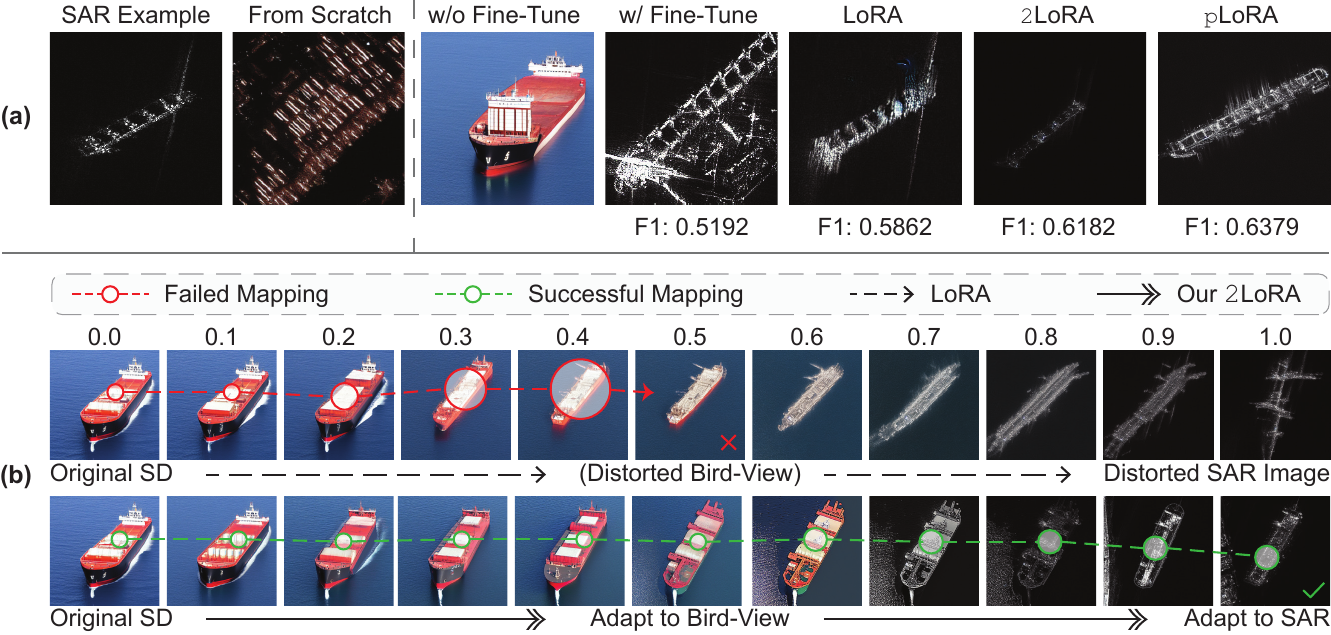}
    \caption{SAR imagery synthesis for ship class ``tanker'', by different methods in (a), and detailed comparison between LoRA and our \texttt{2}LoRA in (b). 
    Methods in (a) include: training an LDM model with SAR data from scratch; inference with a frozen SD; fine-tuning SD; learning a LoRA with SAR data; our \texttt{2}LoRA; our \texttt{p}LoRA. 
    The F1 score of the ``tanker'' ship is given under each column, as a quantitative reference for the utility of synthesized images. In (b), the scale of applying LoRA (or \texttt{2}LoRA) on SD is given on top of each column. Please note that all images are generated with the same text prompt ``A SAR image of a tanker ship".
    Please zoom in for a better view.}
    \label{fig:teaser}
    \vspace{-0.3em}
\end{figure*}

Large-scale pre-trained generative
models, such as Stable Diffusion (SD)~\citep{Rombach2021HighResolutionIS}, Imagen~\citep{Saharia2022PhotorealisticTD}, GLIDE~\citep{Nichol2021GLIDETP}, and ControlNet~\cite{Zhang2023AddingCC} (based on SD) can generate realistic and diverse regular images given textual or structural prompts. 
Their success comes from the capacity of learning a robust and generalizable representation space from billions of web images~\citep{Schuhmann2022LAION5BAO,Kwon2022DiffusionMA}.
In this paper, we focus on synthesizing data for an uncommon data modality Synthetic Aperture Radar (SAR) whose spectrum is outside human's visible light range ($400$-$700$ nm), and has wide applications in marine safety, environmental protection, and climate studies.
In our preliminary study with the open-sourced SD model, we found that fine-tuning it on SAR (whose data is often insufficient) tends to overfit that modality and ``forget'' the general features learned from its original pre-training. 
Another observation is that straightforwardly employing the popular ``anti-forgetting'' domain adaptation method LoRA (low-rank decomposition)~\citep{Hu2021LoRALA} does not solve the problem.
The main challenge is the difficulty in capturing the two primary differences: structure and modality, between regular images and SAR images. 
Fine-tuning or low-rank decomposition on SD fails to capture the necessary feature transformations arising from these two differences.
For example, in Figure~\ref{fig:teaser}~(a), the 4\textsuperscript{th} and 5\textsuperscript{th} images display the synthesis results from the fine-tuned SD and LoRA-based SD, respectively, both of which are suboptimal. The SAR image classification results (F1 scores), using corresponding synthesized images for data augmentation, are given below the images for reference, and the higher the better.

So, the question arises: why, given such a large domain gap, do we still consider using the pre-trained knowledge from regular images? Our motivation comes from an initial empirical observation below. When we don't perform any fine-tuning or LoRA on SD and directly use the SD features (extracted from SAR data) to train a SAR recognition model, we found that this model can easily outperform existing SAR models which are usually small-scale, trained from scratch, or initialized from a model pre-trained on ImageNet~\citep{ridnik2021imagenet}. 
The possible reasons are three-fold. 1)~SAR data contains noise (due to atmospheric conditions, sensor noise, etc.), and the SD model inherently has denoising properties. 2)~SAR data often exhibit high variability (due to factors such as different sensors, acquisition times, and weather conditions), and the SD model based on large-scale pre-training is robust to such variations. 3)~SAR data have spatially coherent structures (\emph{e.g.}, urban areas, water bodies, and forests), and the diffusion process in SD respects spatial coherence, making SD adept at preserving these structures while extracting meaningful features. 
Therefore, we believe if the SD model with its general vision knowledge can be properly adapted to the SAR domain, \emph{e.g.}, to synthesize diverse and high-quality SAR images, it will address the data scarcity issue of SAR.

As aforementioned, directly adapting SD with SAR images is not working due to the large domain gap.
We dive into the detailed visualization of failure cases of LoRA in the 1\textsuperscript{st} row of Figure~\ref{fig:teaser}~(b), taking images generated with prompt ``tanker ship'' as examples.
A LoRA scale of 0 implies the absence of LoRA (mean, original SD), producing 
a regular RGB image of a ``tanker ship". 
As the scale adjusts to 0.5, noticeable distortions emerge, rendering a blurry deck. With a further increase to a scale of 1.0, the distortion becomes egregious, resulting in the very wrong structure of ``tanker ship''.
So, the finding is that the transition from a regular view (scale = 0) to an aerial view (approximately at scale = 0.5) fails in SD, and this failure propagates to result in a flawed SAR image (when the scale reaches 1.0).
An intuitive solution is that enhancing the SD's synthesis ability on aerial-view images may prevent the distortion at the first stage (\emph{i.e.}, from a regular view to an aerial view). It is also a soundable solution since one of the key differences between regular images and SAR images is the structure which is fortunately not significantly different between aerial-view images and SAR images. More fortunately, we can leverage optical remote-sensing (ORS) images whose large-scale datasets~\citep{zhang2021shiprsimagenet,ding2021object,Shi2021ObjectLevelSS} are available, to achieve it.

We thus propose a \underline{2}-stage LoRA\footnote{We don't consider fine-tuning SD's own parameters to avoid catastrophic forgetting of SD's pre-trained knowledge.} approach to adapt SD from regular imagery to SAR imagery indirectly, and we call it \texttt{2}LoRA.
In the first stage, the model is learned to adapt from its regular view to the aerial view without changing its data modality, by training an ORS LoRA module on ORS data.
In the second stage, the model is further adapted from RGB modality to SAR modality, by training a SAR LoRA module on SAR data.
Particularly, in the second stage, we introduce a novel \underline{p}rototype LoRA, dubbed \texttt{p}LoRA, as an improved version of \texttt{2}LoRA, to resolve the class imbalance problem in SAR datasets (\textit{e.g.}, ``tanker ship" in SAR ship classification dataset, and ``road" in SAR semantic segmentation dataset are all minor classes with very limited samples).
For \texttt{p}LoRA, we first cluster all training samples on the feature space regardless of their classes, assuming each cluster captures the primary attributes of a specific SAR imagery prototype, \emph{e.g.}, ``long and slender hull", ``wide deck" or ``angular bow".
Then, we use the samples in each cluster to train an individual \texttt{p}LoRA, and weighted sum all \texttt{p}LoRAs to substitute the vanilla SAR LoRA, where the weights are designed to solve the imbalance problem of prototypes.

Our technical contributions in this paper are two-fold: 1)~a pioneer work of leveraging large-scale pre-trained generative models for synthesizing non-visible light images, \emph{i.e.}, transferring the semantic knowledge pre-learned in regular images to diversify the training data in the SAR domain;
and 2)~a novel \texttt{2}LoRA approach that addresses the domain adaptation challenges from regular images to SAR images, and its improved version \texttt{p}LoRA to solve the class imbalance problem in SAR datasets.
In experiments, we employ the resulting models to synthesize samples for minor classes. This data augmentation, when integrated into the training process of SAR models in image classification as well as segmentation tasks, yields notable improvement.

\section{Related Works}

\noindent\textbf{Data Augmentation with Synthetic Images.} Regular images synthesized by pre-trained diffusion models have been validated to be effective in augmenting regular image datasets~\citep{Zhang2022ExpandingSD, He2022IsSD, Azizi2023SyntheticDF}. One work by~\citet{He2022IsSD} combines large models like GLIDE~\citep{Nichol2021GLIDETP}, T5~\citep{Raffel2019ExploringTL} and CLIP~\cite{Radford2021LearningTV} to generate high-quality regular images. Their methods effectively tackled few-shot, zero-shot, and data imbalance problems in original datasets. 
Another work by~\citet{Rombach2021HighResolutionIS} shows that images generated by Denoising Diffusion Probabilistic Models (DDPMs) can enhance ImageNet pre-trained networks. 
To increase semantic 
variance in generated images, the GIF~\citep{Zhang2022ExpandingSD} is designed by adding a marginal perturbation in the latent space of Latent Diffusion Models (LDMs).
However, all of the existing works generate regular images that share similar visual representations with the pre-training datasets of the large models.
The domain gap between the application and pre-training is small. 
Besides, most of the target datasets (to be augmented) don't really have long-lasting data shortage problems as their images are mainly from daily life or can be collected from ordinary cameras, compared to the non-visible light data on specific objects (\emph{e.g.}, ``ships'' in maritime monitoring) which are hard to collect.
\textbf{Our focus in this work is a non-visible light data synthesis}, taking SAR imagery as a study case. SAR imagery has an inherent large domain gap with regular images regarding both structure and modality. The SAR recognition tasks face long-lasting data scarcity issues due to the military implications.

\noindent\textbf{Conventional Data Augmentation.} Data augmentation methods are widely used to improve the diversity of visual datasets, thereby boosting the generalization ability of trained models~\citep{Shorten2019ASO}. Popular strategies include erasing~\citep{Zhong2017RandomED}, image manipulation~\citep{Wang2017HighResolutionIS}, cutmix~\citep{Yun2019CutMixRS}, and search-based methods~\citep{Cubuk2019AutoAugmentLA,cubuk2020randaugment}. As validated by~\citet{Zhang2022ExpandingSD}, typically operating on existing images with manually specified rules, these strategies are limited to local pixel-wise modifications without introducing novel content or unseen visual concepts.
In comparison, we perform data augmentation from a data synthesis perspective. We introduce a method of leveraging the pre-trained knowledge (\emph{e.g.}, that in SD) to generate various training samples in the SAR domain.

\noindent\textbf{Generative Model Adaptation.}
Adapting pre-trained generative models, such as GAN~\citep{karras2020stylegan,brock2018biggan} and Stable Diffusion~\cite{Rombach2021HighResolutionIS}, aims to transfer the knowledge of these models to synthesize new concepts or out-of-distribution data.
Existing methods can be roughly classified into two categories: \emph{concept-level} and \emph{domain-level}. 
\emph{Concept-level} adaptation modifies the model knowledge to moderate new visual concepts (unseen during pre-training), such as for generating new objects, styles, or certain spatial structures.
Within the GAN paradigm, some works~\citep{ojha2021few-shot-gan,gal2022stylegan-nada} fine-tune the entire generator using regularization techniques. In contrast, some other works try to optimize the crucial part of generator~\citep{kim2022dynagan,alanov2022hyperdomainnet}, or introduce a lightweight attribute adaptor before the frozen generator and a classifier after the frozen discriminator~\citep{yang2021one}.
On top of the SD model,
DreamBooth~\citep{Ruiz2022DreamBoothFT} and Textual Inversion~\citep{Gal2022AnII} try to generate new objects in existing scenes.
ControlNet~\citep{Zhang2023AddingCC} and DragDiffusion~\cite{Shi2023DragDiffusionHD} enable conditional generation from a spatial structure.
HyperDreamBooth~\citep{Ruiz2023HyperDreamBoothHF} customizes the generated image to show a certain human face.
All these adaptation methods are based on the assumption that the ``SD backbone'' could readily generate any unseen concept, \emph{i.e.}, the concept has been seen by SD during its pre-training with regular images. 
\emph{Domain-level} adaptation aims to adapt a pre-trained generative model to a new image domain absent (or very rare) from its pre-training data, such as medical images. 
\cite{Chambon2022AdaptingPV} validate that fine-tuning SD with carefully selected hyperparameters could lead to realistic lung X-ray images. \cite{khader2022medical} find that, compared to GAN, the diffusion model is more capable of encompassing the diversity of medical images.
Unlike these works, \textbf{we focus on the special light domain (SAR) and find that fine-tuning SD fails to capture the needed feature transformation}, due to the significant domain gap between regular and SAR data as well as the data insufficiency of SAR.
\section{Preliminaries}
\label{sec_preliminary}

\begin{figure*}[t]
    \centering
    \includegraphics[width=1\linewidth]{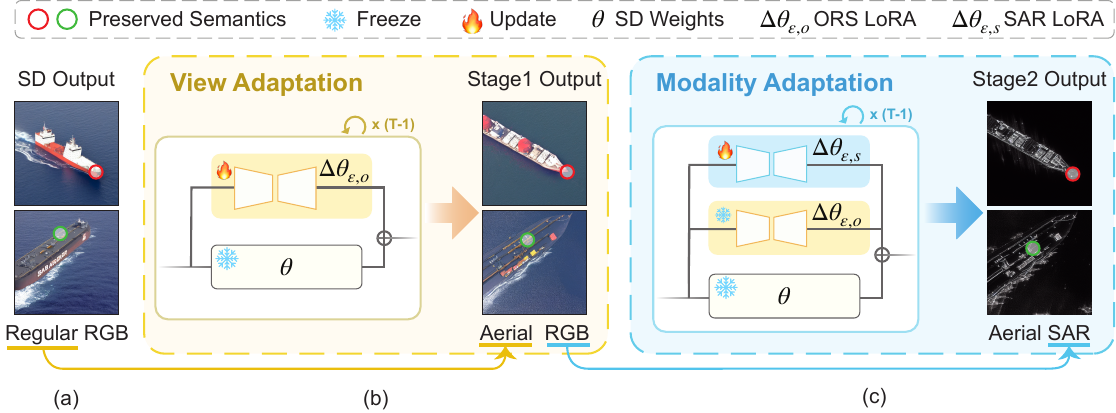}
    \vspace{-2mm}
    \caption{\textbf{The training pipeline of \texttt{2}LoRA.} We train LoRA modules by \texttt{2} stages. 
    In the first stage, \emph{i.e.}, view adaptation, we train an ORS LoRA $\Delta W_\mathrm{o}$ on top of SD's cross-attention layers. 
    In the second stage, \emph{i.e.}, modality adaptation, we train a SAR LoRA $\Delta W_\mathrm{s}$ to further adapt to SAR modality (from RGB modality).
    The learning is on top of a frozen SD-v1.5.
    The ``Regular View" images are generated by SD without any LoRA. 
    The ``Aerial RGB" and ``Aerial SAR" images are the outputs of the two stages of LoRAs, respectively. 
    Particularly, we highlight the preserved semantics, \emph{e.g.}, angle of the bow (in red circles) and yellow oil pipe (in green circles) on the generated images.}
    \vspace{-2mm}
    \label{fig:pipeline}
\end{figure*}

\noindent\textbf{Pre-trained Latent Diffusion Model.} 
We implement our method based on pre-trained latent diffusion model~\citep{Rombach2021HighResolutionIS}. Specifically, we use SD~\cite{runwayml_sd15}.
We chose it because of three aspects: 
1)~SAR images usually suffer from high noise levels, and the SD inherently has a denoising property. 
2)~SAR images often exhibit high variability, and the SD based on large-scale pre-training is robust to such variations.
3)~SAR images have spatial coherent structures, and the diffusion process in SD respects spatial coherence.
Specifically, SD is a text-to-image model incorporating a diffusion process in the latent space of a pre-trained VQGAN autoencoder~\citep{Esser2020TamingTF}. In SD, a denoising U-Net is trained to fit the distribution of latent codes, and it is conditioned on the textual embeddings extracted through a text encoder CLIP~\citep{Radford2021LearningTV} via cross-attention. 
During inference, SD performs iterative reverse diffusion on a randomly sampled noise to generate an image that faithfully adheres to the input text.
Given a data pair $(x,\tau)$\footnote{To avoid confusion with notations in the downstream classification task, we denote text prompts as $\tau$, classification labels as $y$.}, where $x$ is an image and $\tau$ is text prompt, the learning objective for SD is to minimize a denoising objective:
\begin{equation}
\mathcal{L}(x,\tau;\theta) =\mathbb{E}_{\mathcal{E}(x),\tau,\epsilon\sim\mathcal{N}(0,1),t}[\|\epsilon-\epsilon_{\theta}(z_t,t,\psi(\tau))\|^2_2],
\label{eq:SDobj}
\end{equation}
where $z_t$ is the latent feature at timestep $t$, $\psi$ is pre-trained CLIP text encoder, $\mathcal{E}$ is pre-trained VQGAN encoder, and $\epsilon_\theta$ is the denoising U-Net with learnable parameter $\theta$.

\noindent\textbf{Low-Rank Adaptation.}  Our implementation of domain adaptation from regular images to SAR is based on LoRA~\cite{Hu2021LoRALA}. LoRA was initially proposed to adapt large language models to downstream tasks. 
It operates under the assumption that during model updating, parameter updating is sparse. It thus introduces a low-rank factorization of the parameter changes, \textit{i.e.}, $\Delta \theta:=B\small{\cdot} A$. Here, $\theta \in \mathbb{R}^{d\times k}$ represents the parameters of pre-trained model (\emph{e.g.}, SD in our case), and $B\in \mathbb{R}^{d\times r}$ and $A\in \mathbb{R}^{r\times k}$ denote low-rank factors, with $r\ll \min(d,k)$. The updated parameters $\theta'$ are thus given by $\theta'=\theta \small{+} \Delta \theta = \theta \small{+} B\small{\cdot} A$. 
Injecting multiple concepts can be realized by training multiple LoRA modules (each for a single concept) and combining them
through weighted sum,
$
\theta' = \theta + \sum_{i} w_i \Delta \theta_{i}$, and $w_i$ denotes combination weights.

\section{Our Approach}
\label{sec_approach}

Our approach synthesizes high-utility SAR data to augment the minor classes in SAR datasets. We achieve this by adapting SD with the proposed \texttt{2}LoRA and \texttt{p}LoRA.
The \texttt{2}LoRA adaptation has two stages: the view adaptation and the modality adaptation (respectively in Sections~\ref{sec_viewAda} and \ref{sec_modalityAda}). The \texttt{p}LoRA is an improved modality adaptation in the 2\emph{nd} stage, tackling the class imbalance problem of original SAR datasets.
For data augmentation (Section~\ref{sec_aug}), we introduce a method to integrate structural conditioning network (\emph{i.e.}, ControlNet~\citep{Zhao2023UniControlNetAC}) with \texttt{p}LoRA. This allows us to incorporate a wide range of structural conditions from ORS data which further diversifies the synthesized SAR data.

\subsection{View Adaptation}
\label{sec_viewAda}

Figure~\ref{fig:pipeline}~(a) shows that SD without adaptation generates only regular-view RGB images.
To make it adapt from a regular view to an aerial view, in the first stage, we learn an ORS LoRA on top of a frozen SD, as shown in Figure~\ref{fig:pipeline}~(b).

\noindent\textbf{Prompt Construction for ORS}. 
During training, SD associates the visual knowledge in images with the semantics in text prompts, through a cross-attention mechanism.
Low-quality text prompts introduce ambiguity to this association. 
Given the fact that the large-scale ORS datasets~\cite{zhang2021shiprsimagenet,ding2021object,Shi2021ObjectLevelSS} provide only task-specific annotations (\textit{i.e.}, coordinates of bounding boxes or segmentation maps), the question is how to use such annotations to generate high-quality prompts.
Previously, this engineering process was manually done by human prompt engineers. 
In this work, we automate it by leveraging open-sourced Multimodal Large Language Models (MLLMs) such as MiniGPT-4~\cite{zhu2023minigpt}.

Concretely, given a task-specific ORS dataset $\{(x_o, y_o)\}$, we first employ MLLM to identify key visual components (\emph{e.g.}, deck/hull/cabin of ships, or roof/yard of buildings) and contextual factors (\emph{e.g.}, spatial correlations between objects, weather conditions, object orientations) that exhibit distinct visual representations in aerial views. Then, we prompt MLLM to generate detailed text descriptions, which encompass four aspects: spatial correlation between objects; image visibility (\emph{e.g.}, foggy/clear, or dark/bright); orientation of moving objects; visual attributes of key components (\emph{e.g.}, texture, shape, size, and positions).
Finally, we request MLLM to filter out noisy words from these descriptions (\emph{i.e.}, redundant words and personal feelings, such as ``is/and" and ``like/favor") and form a complete prompt for each image. We denote the obtained data triples \{(ORS image, prompt, label)\} as $\{(x_o, \tau_o, y_o)\}$, where the subscript ``o" stands for \underline{O}RS. \emph{We provide chat history with MLLMs and examples of generated prompts in the Appendix.}

\noindent\textbf{ORS LoRA.} We learn an ORS LoRA on the training data $\{(x_o,\tau_o)\}$, to let the SD model ``understand" the aerial view. 
A direct fine-tuning of SD (which is not used in this work) can be formulated with loss $\mathcal{L}(x,\tau;\theta)$ (in Eq.~\ref{eq:SDobj}) as
$\argmin_{\Delta \theta} \mathcal{L}\left(x_o,\tau_o;\theta\small{+}\Delta \theta\right)$,
where $\theta$ denotes parameters of attention layers of denoising U-Net, $(\theta+\Delta \theta)$ denotes updated parameters, ``$+$" means directly modifying $\theta$ with gradient descent, and $\Delta \theta$ indicates the difference. 
We assume this $\Delta \theta$ is low-rank decomposable~\citep{Hu2021LoRALA}, meaning that it can be parameterized by a smaller set of parameters $\varepsilon$, \textit{i.e.}, 
$\Delta \theta:= \varepsilon$.
Therefore, we can search for an optimal low-rank $\varepsilon$ with a parameter-efficient learning objective as:
$\argmin_{\varepsilon} \mathcal{L}\left(x_o,\tau_o;\theta\small{+}\varepsilon \right)$, where $+$ means skip-connecting a layer parameterized by $\varepsilon$ to $\theta$.

Denoising U-Net of SD contains a sequence of attention layers. Hence, to apply the above ORS LoRA to U-Net, we implement $\varepsilon$ by a sequence of lightweight networks, each of them applied to one attention layer of the U-Net, as depicted in Figure~\ref{fig:pipeline}~(b). Each lightweight network is composed of two linear layers with a hidden dimension $r$, i.e., the rank of LoRA.
We apply ORS LoRA networks to only the cross-attention layers (\textit{i.e.}, the Q, K, V linear projection layers), because the research in~\citet{Hu2021LoRALA} has demonstrated that adapting these layers is sufficient and efficient.

We call the learned LoRA networks an ``ORS LoRA'' module, and represent it by $\varepsilon_o$ where ``o" is for \underline{O}RS. 
During the \emph{inference} phase, this module outputs parameter updates $\varepsilon_o$ for each cross-attention layer. The updated network $\theta'$ is thus $\theta' = \theta + w_1\varepsilon_o$,
where $w_1$ is the strength of the module and ``$+$" means skip-connection.

\subsection{Modality Adaptation}
\label{sec_modalityAda}

\noindent\textbf{Prompt Construction for SAR.} 
Due to the inaccurate description of MLLM on SAR modality data, for the SAR prompt, we modify the final step of our prompt construction process in Section~\ref{sec_viewAda}.
We filter out negative prompts that wrongly describe SAR's unique visual features, such as color/texture/shadow/brightness/weather. \emph{Please refer to the Appendix for the full list of negative prompts.} 
We denote the resulting data triples \{(SAR image, prompt, label)\} as $\{(x_s, \tau_s, y_s)\}$, where the subscript ``s" denotes \underline{S}AR.

\begin{figure}[t]
    \centering
    \includegraphics[width=\linewidth]{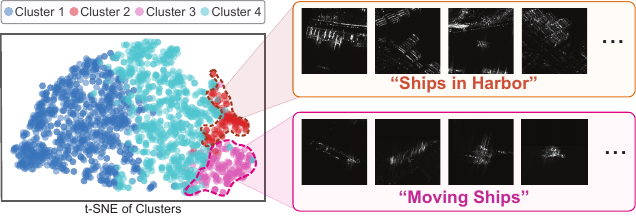}
    \caption{\textbf{Feature clustering results of \texttt{p}LoRA.} The left is the t-SNE feature distribution of the discovered clusters. Each cluster captures an identical visual attribute. On the right, we show samples from two clusters. \textit{Please refer to Appendix for all clusters.}}
    \vspace{-1em}
    \label{fig:cluster}
\end{figure}

\noindent\textbf{\texttt{2}LoRA.} 
In the second stage, we adapt SD \emph{w/} ORS LoRA from RGB to SAR. We achieve this by learning a SAR LoRA on top of SD \emph{w/} ORS LoRA (see Figure~\ref{fig:pipeline}~(c)).
We freeze all modules in SD \emph{w/} ORS LoRA
and train only the SAR LoRA $\varepsilon_s$ with the training data $\{(x_s,\tau_s)\}$. We call the resulting model \texttt{2}LoRA.
During the inference, two LoRAs are weighted combined as:
\begin{equation}
\theta_\mathrm{\texttt{2}LoRA}=\theta+w_1\varepsilon_o+w_2\varepsilon_s
\label{eq:weight_fusion_2lora}
\end{equation}
where $w_1$ and $w_2$ are hyperparameters denoting the strengths of adaptation. 
Note that the sum of $w_1$ and $w_2$ is not necessarily to be $1$. 
As pointed by~\citet{Hu2021LoRALA}, a higher LoRA strength approximates a higher learning rate. In our experiments, we empirically allow $0.5<w_1\small{+}w_2<2$, as we observed clear image distortion 
beyond this range.

\noindent\textbf{\texttt{p}LoRA.} 
As shown in Figure~\ref{fig:teaser}~(b), \texttt{2}LoRA can successfully adapt SD from the regular image domain to the SAR domain. However,  
we observe bias problems from the overall results of \texttt{2}LoRA due to the data imbalance in original SAR datasets (even after resampling). Classes with more samples get higher generation quality.
We tackle this by clustering the SAR dataset into several groups
each representing a visual prototype/attribute of SAR images.
Each prototype contains samples from both major and minor classes, thus reducing the bias.

We first extract the features from SAR images $\{x_s\}$ with a pre-trained SAR
classifier $f$: $\mathbf{F} = f(x)$, where $\mathbf{F}$ denote the features. Then, we apply $K$-Means clustering on $\mathbf{F}$ to get $K$ clusters.
As shown in Figure~\ref{fig:cluster}(a), 
each cluster captures an identical visual attribute such as ``in the harbor" or ``moving".
We split the training data into $K$ groups based on the clusters. Each group is then used to train a prototype LoRA.
We thus get $K$ \underline{p}rototype LoRAs (dubbed \texttt{p}LoRA), and denote their parameters as $\varepsilon_p$, $p \in [1, K]$.

In the \emph{inference} phase, we combine ORS LoRA $\varepsilon_o$ and \texttt{p}LoRA $\varepsilon_p$ vis weighted sum:
\begin{equation}
\theta_\mathrm{\texttt{p}LoRA} = \theta +  w_o\varepsilon_o + \sum^K_{p=1} w_p\varepsilon_p,
\label{eq:weight_fusion_plora}
\end{equation}
where $w_p$ are the weights of prototype LoRAs. Intuitively, this weight $\mathbf{w}\small{=}\{w_p\}$ should be calculated to avoid bias (within a cluster) towards the major classes.
To this end, we calculate the data distribution in each cluster. If a target class (\emph{i.e.}, minor class to be generated) is well-represented in a cluster, we give more weight to that cluster.

Given the cluster distribution $N(C,P)$ where $C=\{c_i\}$ stand for classes, and $P=\{p_j\}$ for clusters, when generating image for minor class $c_i$, the proportion of $c_i$ in each cluster is calculated as 
$b_{i,j}=N(c_i,p_j)/\sum_i N(c_i,p_j)$. We compute a bias vector for minor class $c_i$ as $\mathbf{b}=[b_{i,1},b_{i,2}, ...b_{i,K}]$.
We take these bias vectors as the weights of \texttt{p}LoRA. To ensure the weights sum to $1$, we apply the L1-normalization over $\mathbf{b}$ to get our final \texttt{p}LoRA weights, \textit{i.e.}, $\mathbf{w}$=$\frac{\mathbf{b}}{||\mathbf{b}||_1}$.
\emph{Please kindly refer to the Appendix for more details of computing \texttt{p}LoRA weights and the detailed justification of the advantage of \texttt{p}LoRA over \texttt{2}LoRA.}

\subsection{Dataset Augmentation}
\label{sec_aug}

Our SAR image synthesis method can generate novel data to augment various SAR datasets. We formulate this dataset augmentation process below.
Given an imbalanced SAR dataset of $\mathcal{D}\small{=}\{(x, y)\}$, where the minor classes have only few training samples,
we synthesize new training samples, \textit{i.e.}, $\mathcal{D}'\small{=}\{(x', y)\}$, such that models trained with $\mathcal{D} \cup \mathcal{D}'$ has improved performance, especially for minor classes.

Our principles for data augmentation include 1)~\emph{consistency}: the data distribution in synthetic dataset $\mathcal{D}'$ should align with the original dataset $\mathcal{D}$ in the semantic feature space, \textit{e.g.}, the identical object in real and synthesized images have consistent features such as the structures in SAR domains, and 2)~\emph{diversity}: $\mathcal{D}'$ should offer additional semantic diversity over the existing $\mathcal{D}$, \textit{e.g.}, generating new contexts or novel variants that can enrich existing data.

\noindent\textbf{Consistency.}
To keep consistency, we ensure the structure coherence between real and synthetic images by using structural conditioning networks (\textit{e.g.}, ControlNet~\citep{Zhang2023AddingCC} or T2I Adaptor~\citep{mou2023t2i}).
These networks constrain the generated images to faithfully follow a fixed spatial structure.
Take ControlNet as an example. It learns contour-to-image mapping through a shadow copy of the denoising U-Net, and adds this mapping as a plug-in to SD. 
During inference, given a reference triple (SAR image, prompt, label) denoted as  $(x_s,\tau_s, y_s)$ (constructed in Section~\ref{sec_modalityAda}), we extract structure $c_s$ from $x_s$ (where $c_s$ could be a canny edge, estimated depth, segmentation map, or a stretch
, and feed $c_s$ into ControlNet $\mathcal{C}$. 
The generated image is:
\begin{equation}
x'_s=\mathcal{D}\left(\epsilon_\theta\left(t, \tau_s, \mathcal{C}(c_s)\right)\right),
\label{eq:controlnet}
\end{equation}
where $\mathcal{D}$ is decoder of autoencoder, $\epsilon_\theta$ is the denoising U-Net parameterized by $\theta$, $t$ is timestep, and $\mathcal{C}(c_s)$ is embedding of condition $c_s$ output by ControlNet $\mathcal{C}$. The resulting new training data is $\{(x'_s, y_s)\}$.

\noindent\textbf{Diversity.} To achieve a high diversity of generated data, we borrow the rich structure information from large-scale ORS datasets as a condition of the generation.
This is reasonable as there is a structure coherence between ORS and SAR images (\textit{i.e.}, same objects/regions share the same outlines in these two types of data).
To this end, given a reference triple (ORS image, prompt, label) denoted as $(x_o, \tau_o, y_o)$ (constructed in Section~\ref{sec_viewAda}),
we extract ORS structure (\emph{e.g.}, a canny edge) $c_o$ from $x_o$ and feed it into the ControlNet. Similar to Eq.~\ref{eq:controlnet}, the generated image can be formulated as $x'_o=\mathcal{D}\left(\epsilon_\theta\left(t, \tau_o, \mathcal{C}(c_o)\right)\right)$, and the resulting new training data is denoted as $\{(x'_o, y_o)\}$.
Finally, we use all newly generated data $\mathcal{D}'=\{(x'_s, y_s), (x'_o, y_o)\}$ to augment corresponding categories in the SAR dataset, \textit{i.e.}, the dataset after augmentation is $\mathcal{D} \cup \mathcal{D}'$.

\begin{figure}[t]
    \centering
    \includegraphics[width=\linewidth]{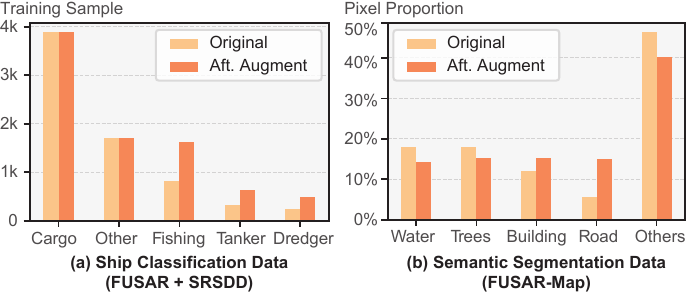}
    \vspace{-2em}
    \caption{\textbf{Data distribution before and after data augmentation.} (a) is the per-class training sample number in the ship classification, and (b) for pixel proportion in the semantic segmentation.}
    \label{fig:dataset}
    \vspace{-1em}
\end{figure}

\section{Experiments}
\label{sec:exp}

We evaluated our SAR synthesis method by augmenting datasets for two typical SAR tasks: fine-grained ship classification and geographic semantic segmentation.

\noindent\textbf{Fine-grained ship classification task.} We construct prompts for \textit{ORS ship data} and \textit{SAR ship data}, and train the generation model on them in two stages (as in \texttt{2}LoRA or \texttt{p}LoRA), respectively. Then, we conduct \textit{dataset augmentation} on the SAR ship dataset for minor classes.
Details are as below.
1) \emph{ORS ship data.} We combined two ORS ship datasets, DOTAv2~\citep{ding2021object} and ShipRS~\cite{zhang2021shiprsimagenet}, to get enough training data.
We use GPT-4~\cite{openai2023gpt4} to construct prompts based on their annotations
(\textit{i.e.}, the \texttt{xml} annotation from ShipRS, and the B-Box coordinates from DOTAv2)
2)~\emph{SAR ship data.} Again to get enough training data, we combined two high-resolution (less then $10$m/pixel) public datasets FUSAR-Ship~\citep{hou2020fusar} and SRSDD~\citep{lei2021srsdd}.
Besides, we merged hard-to-evaluate classes (\emph{i.e.}, classes with fewer than 10 test samples) into the ``others'' class, 
to ensure each preserved class has enough testing samples for stable evaluation of the methods.
\emph{The customized dataset is attached in the Appendix.}
On this dataset, we use GPT-4 to construct text prompts based on their category annotations, following the method in Section~\ref{sec_modalityAda}.
3)~\emph{Dataset augmentation.} As shown by the ``Original" columns in Figure~\ref{fig:dataset}(a), the SAR ship data faces the class imbalance issue. To tackle this problem, we generate new training samples for the minor classes\footnote{Based on the Pareto distribution~\cite{liu2019large,dunford2014pareto} (also known as 20--80 principle), we define categories fewer than 20\% of the total training samples as "minor" classes".} (\textit{i.e.}, the ``fishing", ``tanker", and ``dredger"), following the method described in Section~\ref{sec_aug}.
Specifically, we double the training samples of minor classes, \emph{e.g.}, for ``tanker" ship with $315$ real samples, we synthesize additional $315$ samples.
This dataset distribution change is shown in Figure~\ref{fig:dataset}(a). 
For generating new data, we use the ship outline (\textit{i.e.}, canny edges~\cite{Canny1986ACA} extracted from ORS and SAR ship datasets) as the reference structure $c_s$ and $c_o$ for ControlNet.

\noindent\textbf{Geographic semantic segmentation task.} We use a public dataset \textit{FUSAR-Map}~\cite{Shi2021ObjectLevelSS} to train generation models, and then use these models to perform \textit{dataset augmentation} on the FUSAR-Map dataset for minor classes. 
1)~\emph{FUSAR-Map dataset.} FUSAR-Map~\cite{Shi2021ObjectLevelSS} dataset contains 6,974 training tuples of \{(SAR image, ORS image, segmentation map)\}. The ORS images are sourced from Google Earth and are aligned (via coregistration) with SAR images. 
We use a MiniGPT-v2 (LLaMA-2 Chat 7B)~\cite{chen2023minigpt} to construct prompts for ORS and SAR data, as elaborated in Section~\ref{sec_viewAda} and Section~\ref{sec_modalityAda}.
Especially, with MiniGPT-v2's good understanding of ORS data, the ORS prompts are derived directly from ORS images themselves.
2)~\emph{Dataset augmentation.} We oversampling the minor classes (\emph{i.e.}, the ``building" and ``road") to increase their pixel representation to around 15\%, as shown in Figure~\ref{fig:dataset}(b). During new data generation, we use the segmentation map (from the FUSAR-Map dataset itself) as the reference structure for ControlNet. Especially, to enrich the diversity of segmentation maps, we preserve the target class in them and apply basic transformations (\textit{i.e.}, cut-mix, rotate, random crop) to other classes.

\begin{table}[t]
\centering
\resizebox{\linewidth}{!}{%
\begin{tabular}{@{}lcccccccccc@{}}
\toprule[1.3pt]
\multirow{2}{*}{Method} & \multicolumn{2}{c}{cargo} & \multicolumn{2}{c}{fishing} & \multicolumn{2}{c}{tanker} & \multicolumn{2}{c}{dredger} & \multicolumn{2}{c}{\textbf{Average}}\\ 
\cmidrule(lr){2-3} \cmidrule(lr){4-5} \cmidrule(lr){6-7} \cmidrule(lr){8-9} \cmidrule(l){10-11}
 & FID$_S$$\downarrow$ & F1$\uparrow$ & FID$_S$$\downarrow$ & F1$\uparrow$ & FID$_S$$\downarrow$ & F1$\uparrow$ & FID$_S$$\downarrow$ & F1$\uparrow$  & FID$_S$$\downarrow$ & F1$\uparrow$\\ 
 \midrule \addlinespace[0.05em] \midrule
    \texttt{2}LoRA & 0.848 & \textbf{93.4} & 0.930 & 70.6 & 1.102 & 61.8 & 0.967 & 82.1 & 0.962 & 77.0\\
    \texttt{c}LoRA & \textbf{0.840} & 91.6 & 0.927 & \textbf{72.0} & 1.097 & 60.2 & 0.962 & 81.8 & 0.957 & 76.4 \\
    \texttt{p}LoRA$^\dagger$ & 0.859 & 91.8 & \textbf{0.911} & 71.5 & 0.935 & 62.0 & \textbf{0.912} & 82.5 & \textbf{0.904} & 77.0 \\
    \texttt{p}LoRA & 0.847 & 91.8 & 0.934 & 71.5 & \textbf{0.926} & \textbf{63.8} & 0.939 & \textbf{84.5} & 0.912 & \textbf{77.9}
    \\ \bottomrule[1.3pt]
\end{tabular}
}
\vspace{-1em}
\caption{\textbf{Ablation study.} To address the class imbalance issue, we explored two strategies over \texttt{2}LoRA, namely \texttt{c}LoRA and \texttt{p}LoRA. \texttt{c}LoRA has three \underline{c}ategory-LoRAs respectively trained on minor classes ``fishing", ``tanker", ``dredger"; and \texttt{p}LoRA is four \underline{p}rototype-LoRA as described in Section~\ref{sec_modalityAda}. The ``$\dagger$" denotes using uniform weights $w_p$ (\textit{i.e.}, $w_p=1/4$ for prototype-LoRAs).}
\vspace{-1.5em}
\label{tab:comp_lora}
\end{table}
\begin{table*}[t]
\centering
\resizebox{\linewidth}{!}{%
\rowcolors{2}{white}{white}
\begin{tabular}{lccccccccccccccccc}
\toprule[1.3pt]
\multirow{2}{*}{Method} & \multicolumn{3}{c}{cargo} & \multicolumn{3}{c}{other} & \multicolumn{3}{c}{fishing} & \multicolumn{3}{c}{tanker} & \multicolumn{3}{c}{dredger} & \multicolumn{1}{c}{\multirow{2}{*}{\textbf{AvgF1}}} & \multicolumn{1}{c}{\multirow{2}{*}{\textbf{F1$_m$}}} \\ \cmidrule(lr){2-4} \cmidrule(lr){5-7} \cmidrule(lr){8-10} \cmidrule(lr){11-13} \cmidrule(lr){14-16}
& \multicolumn{1}{c}{Prec.(\%)} & Rec.(\%) & F1(\%) & Prec. & Rec. & F1 & Prec. & Rec. & F1 & Prec. & Rec. & F1 & Prec. & Rec. & F1 \\ \midrule \addlinespace[0.2em] \midrule
AutoAug & 88.71 & 94.69 & 91.60 & 80.28 & 81.02 & 80.65 & 74.60 & 58.39 & 65.51 & 68.00 & 57.63 & 62.39 & 88.10 & 71.15 & 78.72 & 75.77 & 68.87 \\
RandAug & 91.74 & 96.37 & \textbf{94.00} & 81.45 & \textbf{83.33} & \textbf{82.38} & 75.97 & 60.87 & 67.59 & 68.00 & 57.63 & 62.39 & 79.59 & 75.00 & 77.23 & 76.72 & 69.07 \\ 
Oversampling & 89.96 & \textbf{96.37} & 93.06 & \textbf{81.99} & 80.09 & 81.03 & 72.73 & 59.63 & 65.53 & \textbf{69.57} & 54.24 & 60.95 & \textbf{88.64} & 75.00 & 81.25 & 76.36 & 69.24 \\ 
\midrule
Fine-Tune & 86.96 & 95.85 & 91.19 & 80.21 & 71.30 & 75.49 & 72.66 & 57.76 & 64.36 & 60.00 & 45.76 & 51.92 & 88.64 & 75.00 & 81.25 & 72.84 & 65.84 \\
LoRA & 87.96 & 95.60 & 91.62 & 80.22 & 67.59 & 73.37 & 72.66 & 57.76 & 64.36 & 59.65 & 57.63 & 58.62 & 77.78 & 80.77 & 79.25 & 73.44 & 67.41 \\
\rowcolor{gray!16}\texttt{2}LoRA & \textbf{91.85} & 94.95 & 93.38 & 77.88 & 81.48 & 79.64 & 80.80 & 62.73 & 70.63 & 66.67 & 57.63 & 61.82 & 76.67 & 88.46 & 82.14 & \textbf{77.52} & 71.52 \\
\rowcolor{gray!16}\texttt{p}LoRA & 91.72 & 91.84 & 91.78 & 71.13 & 78.70 & 74.73 & \textbf{81.10} & \textbf{63.98} & \textbf{71.53} & 64.91 & \textbf{62.71} & \textbf{63.79} & 76.56 & \textbf{94.23} & \textbf{84.48} & 77.26 & \textbf{73.27} \\ 
\bottomrule[1.3pt]
\end{tabular}
}
\vspace{-0.5em}
\caption{\textbf{Compare dataset augmentation methods for ship classification.} We compare both conventional (rows 1-3) and generative (rows 4-7) data augmentation methods. 
``AutoAug"~\cite{cubuk2019autoaugment} uses an automatic algorithm to search for optimal augmentation strategies. 
``RandAug"~\cite{cubuk2020randaugment} randomly applies basic transformations~\cite{shorten2019survey} to increase the augmentation search efficiency. 
``Oversampling"~\cite{kotsiantis2006handling} doubles the training samples for minor classes.
``Fine-Tune'' uses an SD model fine-tuned on SAR data to generate new data. ``LoRA" uses a single SAR LoRA to generate new data.
Results are presented in Precision (Prec.), Recall (Rec.), and F1-Score (F1) by percentage. The ``AvgF1" is F1-Score averaged on all classes, and F1$_m$ is F1-Score averaged on \underline{m}inor classes. 
Our approaches are marked in \setlength{\fboxsep}{2.5pt}\colorbox{gray!20}{gray}.
}
\label{tab:main_comp}
\end{table*}

\begin{table}[t]
\centering
\resizebox{\linewidth}{!}{%
\begin{tabular}{lccccc}
\toprule[1.3pt]
\multirow{2}{*}{Method}  & \multicolumn{4}{c}{\textbf{Per-class Accuracy} (\%)$\uparrow$} & \multicolumn{1}{c}{\textbf{AvgAcc}} \\ \cmidrule(lr){2-5}
& water & tree & building & road & (\%)$\uparrow$ \\ 
\midrule \addlinespace[0.05em] \midrule
AutoAug & 86.42 & 89.91 & 34.07 & 18.92 & 57.33 \\
RandAug & 87.01 & \textbf{90.07} & 32.19 & 14.92 & 56.05 \\
Oversampling & \textbf{89.41} & 84.31 & 36.58 & 20.19 & 57.62 \\ \midrule
Fine-Tune & 87.06 & 88.25 & 33.49 & 19.42 & 57.06 \\
LoRA & 88.01 & 88.14 & 33.06 & 17.94 & 56.79 \\ 
\rowcolor{gray!16}\texttt{2}LoRA & 87.94 & 88.95 & 34.01 & 19.46 & 57.59 \\
\rowcolor{gray!16}\texttt{p}LoRA & 86.85 & 88.52 & \textbf{37.95} & \textbf{23.49} & \textbf{59.20} \\ 
\bottomrule[1.3pt]
\end{tabular}
}
\vspace{-1em}
\caption{\textbf{Overall comparison on the semantic segmentation task.} Augmentation is performed for two minor classes ``road" and ``building". See Table~\ref{tab:main_comp} caption for the definition of methods.}
\vspace{-1.3em}
\label{tab:comp_ss}
\end{table}
\begin{figure*}[t]
    \centering
   \includegraphics[width=\linewidth]{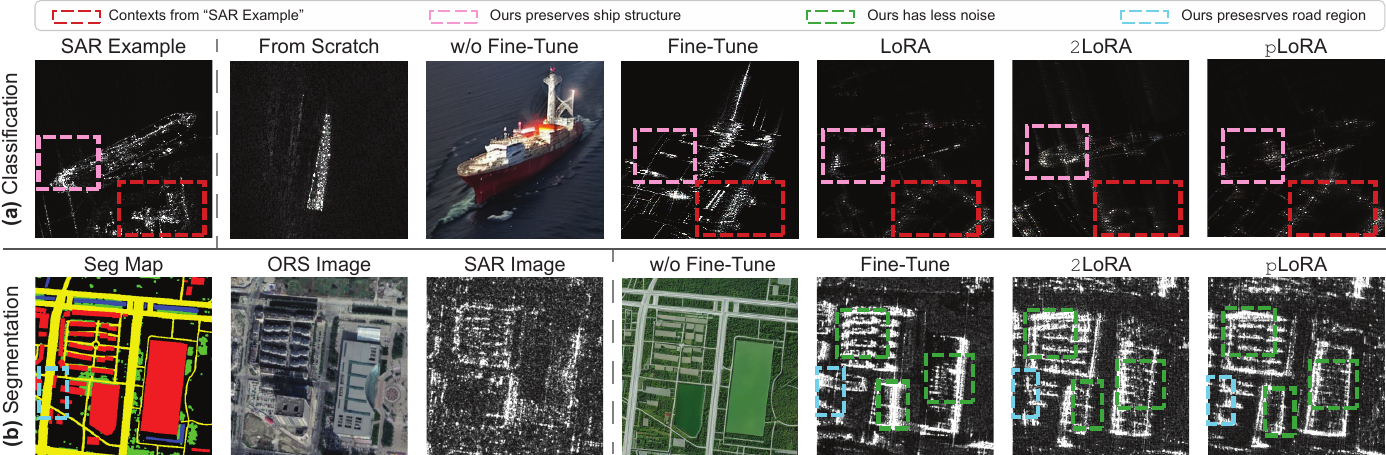}
    \vspace{-2em}
    \caption{\textbf{Qualitative results.} SAR image synthesis for (a) ship classification and (b) geometric semantic segmentation tasks. We show generated samples of minor classes ``fishing" ship in the classification task and ``road'' class in the segmentation task.
    In (a), we extract the Canny edge~\cite{Canny1986ACA} from ``SAR Example" as the reference structure. Please note the content inside the red box is the background.
    }
    \label{fig:qualitative}
\end{figure*}

\noindent\textbf{Evaluation.}
We use a FID$_S$ score (Frechlet Inception Distance for \underline{S}AR) to evaluate image synthesis quality. Unlike conventional FID~\cite{heusel2017gans} which uses an ImageNet pre-trained feature extractor, the FID$_S$ use a FUSAR~\cite{hou2020fusar} pre-trained ResNet50 as the feature extractor. The FID$_S$ is thus customized to SAR data evaluation. It is computed between real and synthetic SAR images. 
All numbers are averaged from the results of five runs of models.
\emph{Please see more implementation and setting details in the Appendix.}

\noindent
\textbf{An ablation study on SAR LoRA.}
When training SAR LoRA on the entire dataset, we see a knowledge bias towards major classes, as in the first row of Table~\ref{tab:comp_lora}.
The FID$_S$ of minor classes are clearly higher than those of major.
To alleviate this issue, we explored two strategies: cLoRA and pLoRA, as described in the caption of Table~\ref{tab:comp_lora}. 
We have three observations from their results in Table~\ref{tab:comp_lora}.
1)~\texttt{c}LoRA has lower (better) FID$_S$ than \texttt{2}LoRA but lower (worse) F1-Score.
This means \texttt{c}LoRA produces similar images to real ones but is not helpful for data augmentation. 
This is because each of its LoRA components is focused on
a specific class without incorporating any new information from other classes, \textit{e.g.}, contexts.
Such focus limits the diversity and utility of the generated images.
2)~Compared to \texttt{2}LoRA and \texttt{c}LoRA, \texttt{p}LoRA exhibits higher F1-Scores in most cases, especially when compared to \texttt{2}LoRA, suggesting its strong capability of alleviating the class imbalance issue and generating high-utility data for augmentation. 
3)~\texttt{p}LoRA consistently outperforms \texttt{p}LoRA$^\dagger$, especially for minor classes (+$1.8$ for ``tanker" and +$2.0$ for ``dredger"). This is because \texttt{p}LoRA is using a more balanced weighting mechanism (in the last paragraph of Section~\ref{sec_modalityAda}) that mitigates the knowledge bias (toward major classes) within every single cluster.

\noindent\textbf{Comparing with data augmentation methods.} We compare both conventional and generative data augmentation methods. For conventional methods, we compared with the state-of-the-art AutoAug~\cite{cubuk2019autoaugment}, RandAug~\cite{cubuk2020randaugment}, and Oversampling~\cite{kotsiantis2006handling}. 
For generative methods, we show the results of 1) fine-tuning SD using SAR data (denoted as ``Fine-Tune''), 2) adapting SD to SAR data using a single LoRA (denoted as ``LoRA''), 3) adapting SD using our proposed \texttt{2}LoRA and 4) \texttt{p}LoRA. 
These methods are further explained in the caption of Table~\ref{tab:main_comp}.

From Table~\ref{tab:main_comp}, we have the following observations. 1)~\texttt{2}LoRA consistently outperforms the methods learned solely on SAR data (\textit{i.e.}, ``Fine-Tuning'' and ``LoRA''), suggesting our view adaptation approach is effective.
2)~Compared with \texttt{2}LoRA, \texttt{p}LoRA improves the classification accuracy of minor classes significantly (+$1.75$ in F1$_m$) at an expense of slightly decreased overall performance (-$0.26$ in AvgF1). 
In this long-tailed recognition setting, it is a commonly observed problem called ``trade-off between head and tail performances''~\cite{li2020overcoming,wang2021adaptive}.
Our future work will explore more to mitigate this trade-off (\textit{e.g.}, by augmenting all classes).
However, in this work, we aim to improve minor classes (by generating data only for minor classes). 
3)~The generative methods are more effective than conventional methods. A simple oversampling readily outperforms RandAug by $0.17$ points in F1$_m$, and our pLoRA further outperforms oversampling (+$4.03$ in F1$_m$). This is because conventional methods focus on batch augmentation (which is dominated by major classes), while ours is specifically designed for minor classes.
We provide the results on \textit{geographic semantic segmentation task} in Table~\ref{tab:comp_ss}. 
We show that \texttt{p}LoRA achieves the best accuracy for ``building" and ``road", and outperforms conventional augmentation methods significantly (+$1.58$ AvgAcc). It is worth mentioning that the \texttt{2}LoRA here is less pronounced due to SD's inherent knowledge about urban-scale remote sensing images (as shown in the ``w/o FineTune" column in Figure~\ref{fig:qualitative}(b)).
We also find that our method is complementary to conventional augmentation methods, \textit{and show evidence in the Appendix.}

\noindent\textbf{Qualitative results.} 
Similar to Figure~\ref{fig:teaser}~(a), we show synthetic examples of minor classes in Figure~\ref{fig:qualitative}. Some key observations include 1) our \texttt{2}LoRA and \texttt{p}LoRA preserves image structures. For example, our method generates the correct SAR representation for ship components (pink boxes);  our method preserves ``road" regions correctly (blue boxes); 2) our method presents less noise and clearer details (green boxes). \emph{More visual examples are given in the Appendix.}

\section{Conclusions}
We explored the hidden potentials of large-scale pre-trained image generation models in non-visible light domains. We revealed some intriguing findings that led us to the 2-stage low-rank adaptation method \texttt{2}LoRA and its improved version \texttt{p}LoAR. 
Future research could include the applications of our method in more non-visible light domains such as MRI and infrared imaging.

\section*{Acknowledgements}
The author gratefully acknowledges the support of the DSO research grant awarded by DSO National Laboratories (Singapore), and the Lee Kong Chian Fellow grant awarded to Dr Qianru Sun by Singapore Management University.

{
    \small
    \bibliographystyle{ieeenat_fullname}
    \bibliography{main}
}

\clearpage
\newpage

\setcounter{table}{0}
\renewcommand{\thetable}{S\arabic{table}}
\setcounter{figure}{0}
\renewcommand{\thefigure}{S\arabic{figure}}
\setcounter{section}{0}
\renewcommand{\thesection}{\Alph{section}}

\noindent
{\Large {\textbf{Supplementary materials}}}
\\

\noindent This appendix is organized as follows:

\begin{itemize}
    \item Section~\ref{sec_just} justifies the weight $w_p$ used in \plora (Eq.~(3)).
    \item Section~\ref{sec_impl} provides implementation details of comparison experiments. We also elaborate on the data pre-processing details. 
    \item Section~\ref{sec_exps} shows additional experimental results and findings, including:
    \begin{itemize}[label=\textopenbullet]
        \item Combination of our method with conventional data augmentation methods (line~574 in Section~5).
        \item Visualization of cluster samples (Figure~3).
        \item Visualization of our synthetic images (line~584 in Section~5).
        \item Impact of different LoRA ranks.
        \item Using a smaller training dataset.
    \end{itemize}
\end{itemize}

\section{Justification}
\label{sec_just}

\noindent\textbf{Justification of weight $w_p$ in Eq.~(3).} We justify that the combination weight $w_p$ makes \plora less prone to data bias problems then \twolora or LoRA. Take the ``fishing" ship generation as an example: 

1) In \texttt{2}LoRA or LoRA, we use the entire training dataset (see Table~\ref{tab:4attrilora} last row), where ``fishing" ships represent only $13.4\%$. Here, dominant classes like ``cargo" ($49.4\%$) overshadow the ``fishing" class.

2) In \texttt{p}LoRA, alternatively, we use four feature clusters (rows 1-4 in Table~\ref{tab:4attrilora}) to train four prototype LoRAs. For the "fishing" ship class, take clusters $p_3$ and $p_4$ as examples: cluster $p_4$ contains a higher proportion ($34.2\%$ in row 3), while $p_3$ has less ($7.9\%$ in row 4). 
Thus, ``fishing" receives a higher weight in $p_4$ (\textit{i.e.}, $w_4=\frac{34.2\%}{11.2\%+13.7\%+7.9\%+34.2\%}=0.51$), and a lower weight in $p_3$ (\textit{i.e.}, $w_3=\frac{7.9\%}{11.2\%+13.7\%+7.9\%+34.2\%}=0.11$). Compared with \texttt{2}LoRA, by using this weight $w_p$, our \plora avoids over-reliance on biased knowledge.

\begin{table}[h]
\centering
\resizebox{\linewidth}{!}{%
\begin{tabular}{@{}cccccc@{}}
\toprule[1.3pt]
\multirow{2}{*}{Clusters} & \multicolumn{5}{c}{Categories} \\
\cmidrule(l){2-6}
& cargo & other & fishing & tanker & dredger \\ \midrule \addlinespace[0.01em] \midrule
All & 49.4\% & 28.1\% & 13.4\% & 5.2\% & 4.0\% \\ \midrule 
$p_1$ & 48.0\% & 27.8\% & 11.2\% & 7.6\% & 5.4\% \\
$p_2$ & 48.3\% & 31.3\% & 13.7\% & 3.3\% & 3.4\% \\
$p_3$ & 73.3\% & 13.4\% & 7.9\% & 4.0\% & 1.4\% \\
$p_4$ & 35.1\% & 29.4\% & 34.2\% & 0.9\% & 0.5\% \\ \bottomrule[1.3pt]
\end{tabular}
}
\caption{\textbf{Category distribution across clusters.} This table shows the proportion of different categories in each cluster $p_k$, and each row sums to $100\%$. ``All" means all of the training data, and $p_1$ to $p_4$ are four clusters from feature clustering.}
\vspace{-1em}
\label{tab:4attrilora}
\end{table}

\section{Implementation Details}
\label{sec_impl}

\subsection{Implementation Details of Table~2}
\label{sec:tab2}
In Table~2, we evaluated our SAR synthesis method by augmenting datasets for fine-grained ship classification tasks. We provide implementation details as follows.

\noindent\textbf{Details of compared methods.} 
For AutoAug~\cite{cubuk2019autoaugment} and RandAug~\cite{cubuk2020randaugment}, we adhere to their original image transformation protocols (including contrast/brightness/sharpness enhancement, rotation, flipping, colour jitting, and shearing, as specified in ~\cite{mmcls_augpolicies}). 
For Oversampling~\cite{kotsiantis2006handling}, we double the training samples of minor classes, \emph{e.g.}, for the ``tanker" ship with $315$ real images, we re-sample an additional $315$ images from its training data.

\noindent\textbf{Details of \plora and \twolora.} We train ORS LoRA for 200 epochs, and train SAR LoRAs (including prototype LoRAs) for 100 epochs. We use the cosine annealing scheduler starting with the learning rate of $1 \times 10^{-3}$ (with batch size $128$). We fine-tune the ControlNet for $30$ epochs using a learning rate of $4 \times 10^{-5}$ (with batch size $4$). 
During inference (image generation), we use the original inference configs of CannyEdge ControlNet~\cite{Zhang2023AddingCC,mikubill_sdwebuicontrolnet}, except that we set sampling steps at $50$, and a combination weight (with SD) at $0.4$.
When training the ship classifier, we use an ImageNet-21K pre-trained ResNet50 as the backbone and fine-tune it on SAR data for 100 epochs. The scheduler here also follows cosine annealing but includes a warm-up phase lasting for about 500 iterations (with batch size $128$). We use the SGD optimizer with a learning rate of $0.1$, momentum of $0.9$, and a weight decay of $1 \times 10^{-4}$.

\noindent\textbf{Evaluation.} To ensure a fair and robust evaluation, we conduct our evaluation as follows. 1) For FID$_S$ score evaluation, we use only the SAR ship contours to ensure a consistent real-to-synthetic image pairing. 2) In generating new training samples, all generative methods (\textit{i.e.}, LoRA, \twolora and \texttt{p}LoRA) are conditioned using the same set of contours. 3) For F1-Score evaluation, we average the results over five runs to enhance evaluation reliability.

\subsection{Implementation Details of Table~3}
In Table~3, we evaluated our SAR synthesis method by augmenting datasets for geographic semantic segmentation tasks. We provide implementation details as follows.

\noindent\textbf{Details of \plora and \twolora.} For LoRA training, we use the same setup as in Section~\ref{sec:tab2}. For ControlNet, we use semantic segmentation maps (from the FUSAR-Map dataset itself) as conditions and train from scratch for $10000$ steps with a batch size of $16$ (\textit{i.e.}, approximate $65$ epochs), with a combination weight (with SD) of $1.0$. During inference, unlike the classification task that uses per-class sample number proportion to calculate weight $w_p$, we here use per-class pixel proportion to calculate weight $w_p$.
\begin{figure*}[t]
    \centering
    \includegraphics[width=1\linewidth]{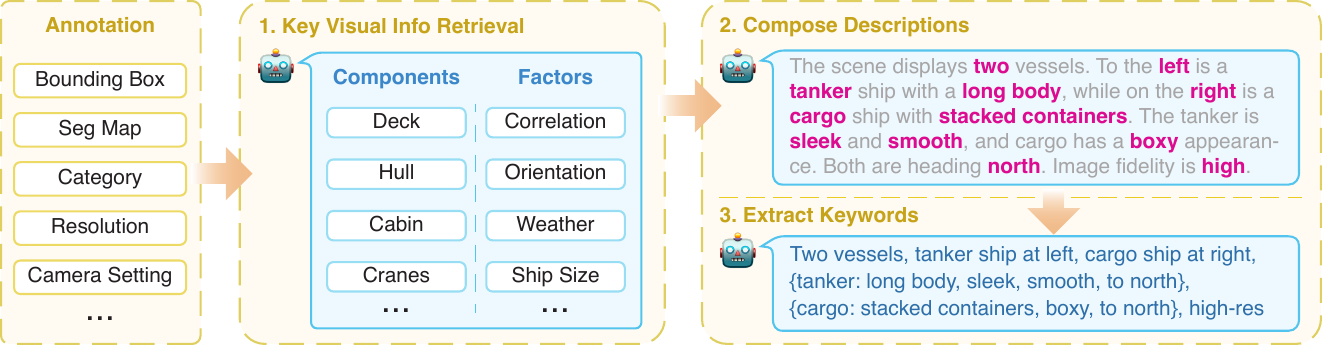}
    \caption{\textbf{Prompt construction workflow.} This figure is to facilitate understanding of our prompt construction workflow in Section~4.1. We construct high-quality prompts by using Multimodal Large Language Models (MLLMs, such as GPT-4 and MiniGPT-v2) in three steps: 1) MLLMs first identify essential visual components and contextual factors from the dataset, and then 2) generate comprehensive descriptions covering spatial relationships, visibility conditions, object orientations, and component attributes. 3) Finally, these descriptions are condensed into keyword-based prompts. The robot emoji (\raisebox{-1pt}{\includegraphics[height=1em]{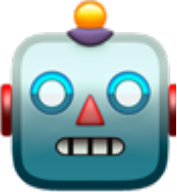}}) indicates the MLLMs.}
    \label{fig:chat}
\end{figure*}

\noindent\textbf{Details of compared methods.} The implementation of AutoAug and RandAug remains the same as in Section~\ref{sec:tab2}. For Oversampling, we only increase the pixel representation for minor classes (\textit{i.e.}, the ``road" and ``building"). Concretely, we selected the top $20\%$ of images with the highest pixel proportions for each category (for instance, from $10,000$ images containing the ``road'' class, the top $2,000$ were chosen). These images are then resampled until their corresponding classes' pixel proportions reach approximately $15\%$, as illustrated in Figure~4(b). \textit{Please note that our generative methods also use this selected set of images to create new samples, as introduced next.}

\subsection{Details of Data Processing}

\begin{figure}[t]
    \centering
    \includegraphics[width=\linewidth]{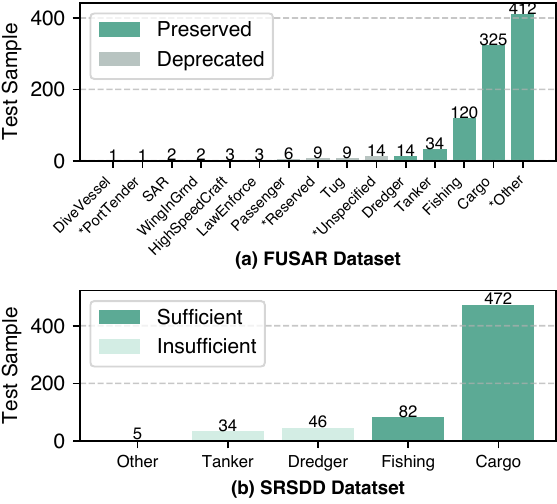}
    \caption{\textbf{Test-set limitations of the SAR datasets.} This figure shows the per-category test sample distribution of the FUSAR dataset (in (a)) and the SRSDD dataset (in (b)). We can observe that the FUSAR dataset faces problems of insufficient test samples and vaguely defined classes (indicated by ``$\ast$"). Besides, the SRSDD dataset also has the problem of insufficient test samples.}
    \label{fig:fusar}
\end{figure}

\noindent \textbf{Details of SAR ship dataset.} We choose FUSAR-Ship~\cite{hou2020fusar} and SRSDD~\cite{lei2021srsdd} datasets as our source datasets because they are high-resolution ($\leq 10$m) and have fine-grained subcategories of ships, as shown in Table~\ref{tab:sar-dataset-list}.
However, both datasets have limitations. As shown in Figure~\ref{fig:fusar}(a), the FUSAR-Ship dataset has insufficient test samples (\textit{i.e.}, certain categories have only $\leq15$ test samples), unclear category definition (\textit{e.g.}, ``Reserved" or ``Unspecified" categories). As shown in Figure~\ref{fig:fusar}(b), the SRSDD dataset has insufficient test samples. To address these issues and establish a robust benchmark,
we combined the ship categories from both
datasets, omitting those with fewer than 10 test samples into an ``others'' category. We show the data distribution of the resulting dataset in Table~\ref{tab:fusrs}. \textit{This dataset is available at this \href{https://github.com/doem97/gen_sar_plora}{GitHub link}.}
\begin{table}[t]
\centering
\resizebox{\linewidth}{!}{%
\rowcolors{2}{white}{white}
\begin{tabular}{lccrll}
\toprule
Dataset & Year & Category & Instances & Width (px) & Resolution (/px) \\ \midrule
OpenSARShip2~\citep{li2017opensarship}$^\dagger$ & 2017 & 16 & 19,360 & 30--120 & 22m \\
SAR-Ship-Dataset~\citep{wang2019sar}$^\ast$ & 2019 & 1 & 59,535 & 256 & 3m--25m \\
AIR-SARShip-2.0~\citep{wang2023saraircraft}$^\ast$ & 2020 & 1 & 461 & 1000 & 1m, 3m \\
\rowcolor{gray!16}FUSAR-Ship~\citep{hou2020fusar} & 2020 & 15 & 6,358 & 512 & $\geq$0.5m \\
HRSID~\citep{wei2020hrsid}$^\ast$ & 2020 & 1 & 16,951 & 800 & 0.5m, 1m, 3m \\
LS-SSDD-v1.0~\citep{zhang2020ls}$^\ast$ & 2020 & 1 & 6,015 & 16,000 & 20m \\
Official SSDD~\citep{zhang2021sar}$^\ast$ & 2021 & 1 & 2,456 & 190--160 & 1m--15m \\
\rowcolor{gray!16}SRSDD-v1.0~\citep{lei2021srsdd} & 2021 & 6 & 2,884 & 512 & 1m \\
RSDD-SAR~\citep{congan2022rsdd}$^\ast$ & 2022 & 1 & 10,263 & 512 & 2m--20m \\
xView3-SAR~\citep{paolo2022xview3}$^\dagger$ & 2023 & 2 & 243,018 & 512 & 10m \\ \bottomrule
\end{tabular}
}
\caption{\textbf{Review of SAR ship datasets.} Only FUSAR-Ship and SRSDD-v1.0 (marked in \setlength{\fboxsep}{2.5pt}\colorbox{gray!20}{gray}) meet our dataset criteria. Datasets marked by ``$\dagger$" means it is deprecated due to \textit{low resolution}, and ``$\ast$" means \textit{insufficient ``ship" subcategories}.}
\label{tab:sar-dataset-list}
\end{table}
\begin{table}[h]
\centering
\resizebox{\linewidth}{!}{%
\begin{tabular}{lrrrrrr}
\toprule[1.3pt]
\multirow{2}{*}{Split} & \multicolumn{5}{c}{Category} & \multirow{2}{*}{Total} \\
\cmidrule(lr){2-6}
& cargo & other & fishing & tanker & dredger & \\ \midrule \addlinespace[0.01em] \midrule
Train & 3,890 & 1,710 & 814 & 315 & 242 & 6,971 \\
Deci & 389 & 171 & 81 & 32 & 24 & 696 \\
Val & 484 & 213 & 102 & 40 & 30 & 869 \\
Test & 772 & 216 & 161 & 59 & 52 & 1,260 \\ \midrule
Total & 5,146 & 2,139 & 1,077 & 414 & 324 & 9,100 \\ \bottomrule[1.3pt]
\end{tabular}
}
\caption{\textbf{Data distribution of our SAR dataset} (based on FUSAR-Ship~\cite{hou2020fusar} and SRSDD-v1.0~\cite{lei2021srsdd}).
Other than conventional ``Train", ``Validation", and ``Test" splits, we especially introduce a new ``Deci" (\textit{i.e.}, ``decimal" meaning one-tenth) split, sampled from the training set. We use ``Deci" to show if the proposed method can work given only very limited training samples.}
\label{tab:fusrs}
\end{table}

\noindent \textbf{Details of ORS ship dataset.} 
We combined the DOTAv2~\cite{ding2021object} and the ShipRSImageNet~\cite{zhang2021shiprsimagenet} datasets to build our ORS dataset. Especially, because the DOTAv2 dataset is originally a detection dataset, we crop out the ship/harbour instances (total $50,356$ instances) as image patches.
\textit{This dataset is available at the \href{https://github.com/doem97/gen_sar_plora}{GitHub link}.}

\begin{table*}[t]
\centering
\resizebox{\linewidth}{!}{%
\rowcolors{2}{white}{white}
\begin{tabular}{lccccccccccccccccc}
\toprule[1.3pt]
\multirow{2}{*}{Method} & \multicolumn{3}{c}{cargo} & \multicolumn{3}{c}{other} & \multicolumn{3}{c}{fishing} & \multicolumn{3}{c}{tanker} & \multicolumn{3}{c}{dredger} & \multicolumn{1}{c}{\multirow{2}{*}{\textbf{AvgF1}}} & \multicolumn{1}{c}{\multirow{2}{*}{\textbf{F1$_m$}}} \\ \cmidrule(lr){2-4} \cmidrule(lr){5-7} \cmidrule(lr){8-10} \cmidrule(lr){11-13} \cmidrule(lr){14-16}
& \multicolumn{1}{c}{Prec.(\%)} & Rec.(\%) & F1(\%) & Prec. & Rec. & F1 & Prec. & Rec. & F1 & Prec. & Rec. & F1 & Prec. & Rec. & F1 \\ \midrule \addlinespace[0.2em] \midrule
Oversampling & 89.36 & 92.49 & 90.90 & 70.56 & 75.46 & 72.93 & 65.47 & 56.52 & 60.67 & 40.82 & 33.90 & 37.04 & 64.29 & 51.92 & 57.45 & 63.80 & 51.72 \\ 
\midrule
Fine-Tune & 89.38 & 92.62 & 90.97 & 71.56 & 74.54 & 73.02 & 64.54 & 56.52 & 60.26 & 36.00 & 30.51 & 33.03 & 59.09 & 50.00 & 54.17 & 62.29 & 49.15 \\
\rowcolor{gray!16}\texttt{2}LoRA & 90.58 & \textbf{93.39} & \textbf{91.96} & \textbf{72.53} & \textbf{78.24} & \textbf{75.28} & \textbf{68.12} & \textbf{58.39} & \textbf{62.88} & \textbf{44.00} & 37.29 & 40.37 & \textbf{67.44} & 55.77 & 61.05 & \textbf{66.31} & \textbf{54.77} \\
\rowcolor{gray!16}\texttt{p}LoRA & \textbf{91.11} & 90.28 & 90.70 & 67.21 & 76.85 & 71.71 & 63.57 & 50.93 & 56.55 & 42.37 & \textbf{42.37} & \textbf{42.37} & 60.00 & \textbf{69.23} & \textbf{64.29} & 65.12 & 54.40 \\ 
\bottomrule[1.3pt]
\end{tabular}
}
\caption{\textbf{Compare augmentation methods under decimal training setting.} Our methods are marked in \setlength{\fboxsep}{2.5pt}\colorbox{gray!20}{gray}.
}
\label{tab:deci}
\end{table*}

\noindent\textbf{Prompt construction.} To facilitate understanding of Section~4.1 in the main paper, we illustrate the prompt construction workflow in Figure~\ref{fig:chat}. 
\textit{The detailed prompt construction process (\emph{i.e.}, chat history with GPT-4 or MiniGPTv2) is available in the \href{https://github.com/doem97/gen_sar_plora}{GitHub link}.} 

\begin{table}[t]
\centering
\resizebox{\linewidth}{!}{%
\begin{tabular}{@{}cccccccccccc@{}}
\toprule[1.3pt]
\multirow{2}{*}{Rank} & \multirow{2}{*}{Params (M)} & \multicolumn{2}{c}{cargo} & \multicolumn{2}{c}{fishing} & \multicolumn{2}{c}{tanker} & \multicolumn{2}{c}{dredger} \\ \cmidrule(lr){3-4} \cmidrule(lr){5-6} \cmidrule(lr){7-8} \cmidrule(l){9-10}
 &  & FID$_S$$\downarrow$ & F1$\uparrow$ & FID$_S$$\downarrow$ & F1$\uparrow$ & FID$_S$$\downarrow$ & F1$\uparrow$ & FID$_S$$\downarrow$ & F1$\uparrow$ \\ \midrule\midrule
4 & 0.0797 & 0.9042 & \textbf{0.9351} & 0.9290 & 0.7088 & 1.0714 & \textbf{0.6296} & 1.1296 & 0.8214 \\
8 & 0.1594 & 0.8504 & 0.9338 & \textbf{0.8970} & 0.7092 & 1.1492 & 0.5981 & 1.0079 & 0.8142 \\
16 & 0.3188 & \textbf{0.8479} & 0.9338 & 0.9301 & 0.7063 & \textbf{1.0227} & 0.6182 & \textbf{0.9670} & \textbf{0.8214} \\
32 & 0.6377 & 0.8519 & 0.9344 & 0.9094 & \textbf{0.7138} & 1.0907 & 0.6038 & 1.0714 & 0.8000 \\ \bottomrule[1.3pt]
\end{tabular}
}
\caption{\textbf{Compare ranks of \twolora.} We compare how different LoRA ranks affect the generated image quality in the ``modality adaptation" stage.}
\label{tab:lora_rank_cls}
\end{table}

\begin{table}[t]
\centering
\resizebox{\linewidth}{!}{%
\begin{tabular}{lccccc}
\toprule[1.3pt]
Method & cargo & fishing & tanker & dredger & \textbf{AvgF1} \\ \midrule\midrule
Augmented dataset & 91.78 & 71.53 & 63.79 & 84.48 & 77.90 \\
+ Oversampling & 91.60 & \textbf{72.29} & \textbf{64.95} & 82.96 & 77.95 \\
+ RandAug & \textbf{92.46} & 72.10 & 64.82 & \textbf{85.26} & \textbf{78.66} \\ \bottomrule[1.3pt]
\end{tabular}
}
\caption{\textbf{Compatibility of our method.} The ``Augmented dataset" refers to the dataset reported as ``\plora" in Table 2 of the main paper. We applied conventional augmentation methods (\textit{i.e.}, Oversampling~\cite{kotsiantis2006handling} and RandAug~\cite{cubuk2020randaugment}) to this dataset. For Oversampling, we re-sample the minor classes to make their training sample number align with the largest category, \textit{i.e.}, cargo. Results are reported in F1-Score and Average F1-Score (AvgF1).}
\label{tab:our+oversample}
\end{table}

\begin{figure}[t]
    \centering
    \includegraphics[width=\linewidth]{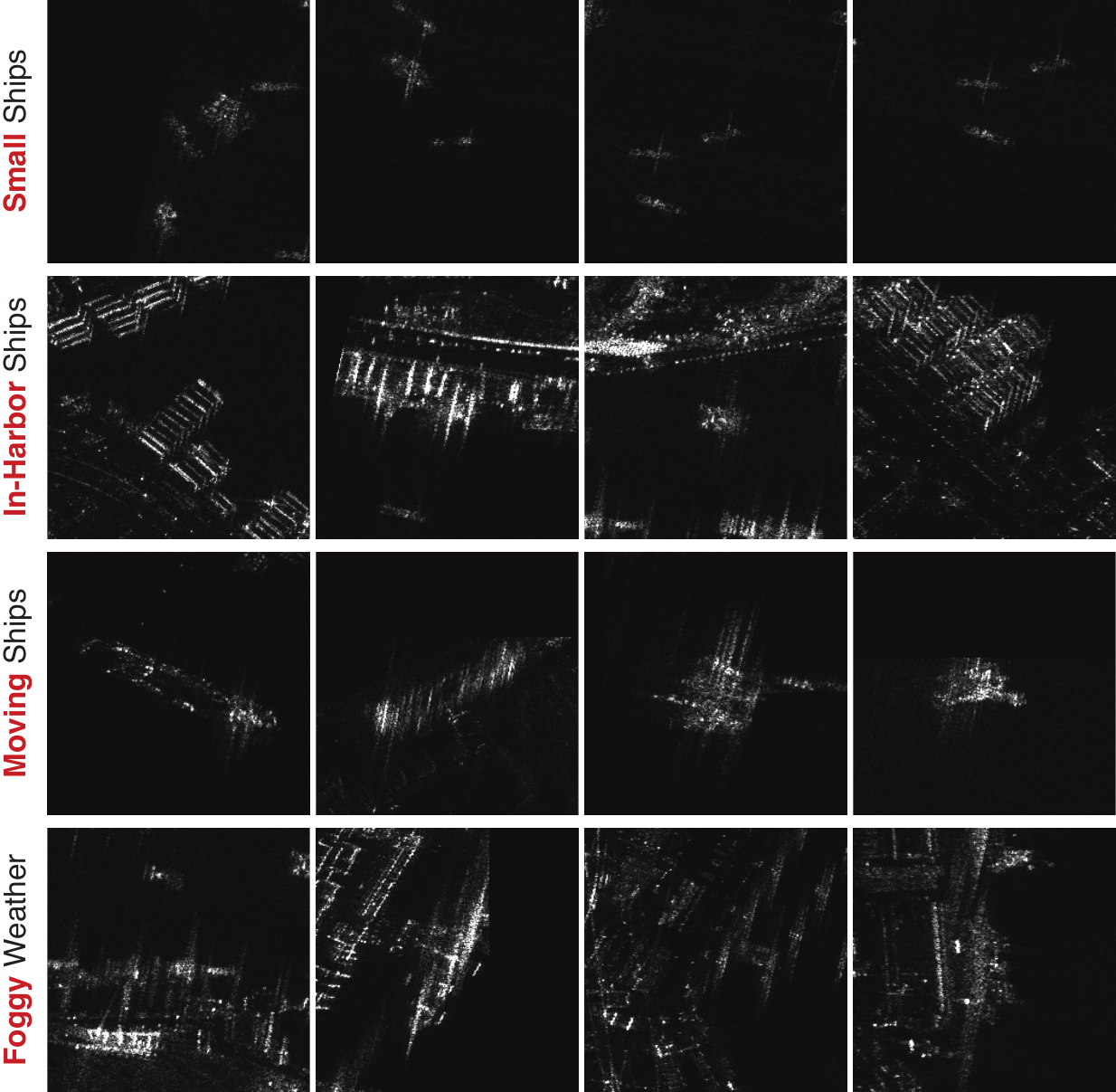}
    \caption{\textbf{Illustration of obtained clusters.} As a supplement to Figure 3, we display more samples from four obtained clusters. Each cluster captures distinct visual representations, such as ``fast-moving", ``small" ships, or ``foggy" weather. \emph{This supplements the Figure 3 in the main paper.}}
    \vspace{-1em}
    \label{appfig:cluster}
\end{figure}

\section{Additional Results}
\label{sec_exps}

\noindent\textbf{Combined with conventional augmentation.} Our method is compatible with conventional data augmentation techniques, as our synthetic images are adaptable to go through a second step of data augmentation using traditional augmentation methods. We show the results of using Oversampling~\cite{kotsiantis2006handling} and RandAug~\cite{cubuk2020randaugment} (as a second step of data augmentation after our method) in Table~\ref{tab:our+oversample}. 

\noindent\textbf{Visualization of more cluster samples (Fig.~3).} After feature clustering, the resulting clusters capture distinct visual representations, such as ship characteristics (\textit{e.g.}, ``fast-moving", ``in the harbor", or ``small-sized") and weather conditions (\textit{e.g.}, ``foggy"). See Figure~\ref{appfig:cluster} for examples. \emph{This supplements the Figure 3 in the main paper.}
\begin{figure*}[t]
    \centering
    \includegraphics[width=\linewidth]{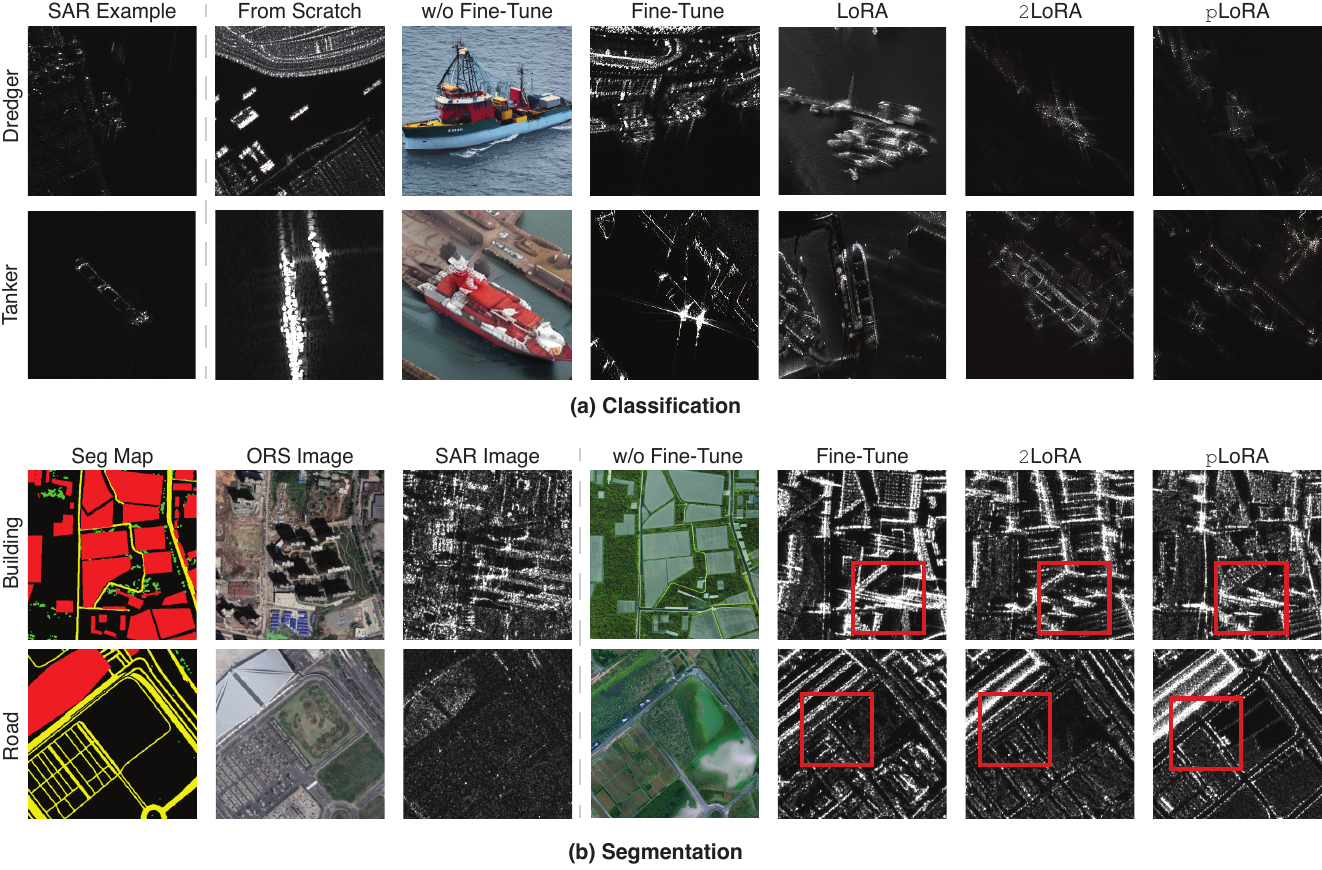}
    \vspace{-2em}
    \caption{\textbf{Qualitative comparison.} This figure demonstrates SAR image synthesis for (a) ship classification and (b) geographic semantic segmentation tasks. Our \plora and \twolora increase pixel-level clarity, significantly reducing noise in synthesized ship instances and segment regions. The red boxes in (b) indicate our methods generate less noise.
    }

    \vspace{-1em}
    \label{appfig:qualitative}
\end{figure*}

\noindent\textbf{Decimal training set.} To define an even more severe data-shortage problem, we split a ``Deci" set, where ``Deci" means Decimal (one-tenth) of the training set, by randomly sampling $1/10$ samples from the whole training data (as shown by ``Deci" row in Table~\ref{tab:fusrs}). We show experimental results by using the models trained on this ``Deci" set in Table~\ref{tab:deci}: \plora and \twolora consistently surpass the other methods. Notably, \plora appears predominantly focused on enhancing the performance of minor classes, whereas \texttt{2}LoRA brings improvements for both minor and major classes.

\noindent\textbf{LoRA ranks.} The rank of the LoRA component controls the number of learnable parameters (\textit{i.e.}, the rank is a hyperparameter). As shown in Table~\ref{tab:lora_rank_cls}, we compare the FID$_S$ and F1-Score of \texttt{2}LoRA generated images with different LoRA ranks. \textit{Please note this is an exploratory study, not for hyperparameter optimization. Hence, we report test set results rather than validation results.}

\noindent\textbf{Visualization of more synthetic images (Fig.5).} Supplementing the Figure~5, we show more qualitative comparison results in Figure~\ref{appfig:qualitative}.

\end{document}